\documentclass[
]{ceurart}

\usepackage{listings}
\usepackage[utf8]{inputenc}
\usepackage[english]{babel}
\usepackage{graphicx, caption}
\usepackage[export]{adjustbox}
\usepackage{svg}
\usepackage{amssymb} 
\usepackage{amsthm} 
\usepackage[onehalfspacing]{setspace}
\usepackage{csquotes}
\usepackage{float}
\usepackage{amsmath}
\usepackage{subcaption}

\begin{document}

  
\conference{BHDC 2023: Big Historical Data Conference, November 22-25, 2023, Jena, Germany}

\title{Trajectories of Change: Approaches for Tracking Knowledge Evolution}

\author[1]{Raphael Schlattmann}[%
email=raphael.schlattmann@tu-berlin.de,
orcid=0000-0002-2234-3720
]
\cormark[1]

\author[2]{Malte Vogl}[%
email=vogl@gea.mpg.de,
orcid=0000-0002-2683-6610
]

\address[1]{Institute of History and Philosophy of Science, Technology, and Literature, Technische Universität Berlin, Germany}
\address[2]{Max Planck Institute of Geoanthropology, Jena, Germany}

\cortext[1]{Corresponding author.}

\begin{keywords}
 quantitative corpus analysis
\sep diachronic variation in language use
\sep evolution of knowledge systems
\sep history of science
\sep history of general relativity
\end{keywords}

\maketitle

\begin{abstract}
We explore local vs. global evolution of knowledge systems through the framework of socio-epistemic networks (SEN), applying two complementary methods to a corpus of scientific texts. The framework comprises three interconnected layers—social, semiotic (material), and semantic—proposing a multilayered approach to understanding structural developments of knowledge. To analyse diachronic changes on the semantic layer, we first use information-theoretic measures based on relative entropy to detect semantic shifts, assess their significance, and identify key driving features. Second, variations in document embedding densities reveal changes in semantic neighbourhoods, tracking how concentration of similar documents increase, remain stable, or disperse. This enables us to trace document trajectories based on content (topics) or metadata (authorship, institution). Case studies of Joseph Silk and Hans-Jürgen Treder illustrate how individual scholar's work aligns with broader disciplinary shifts in general relativity and gravitation research, demonstrating the applications, limitations, and further potential of this approach. 
\end{abstract}

\section{Introduction}
The analysis of the formation and evolution of knowledge systems is of great interest in the history of science. The study of the emergence and development of a scientific field is an example of such an analysis, but is inherently one of structural changes, incorporating shifts in social, material and cognitive configurations. A key challenge is systematically integrating local micro-histories into these broader reconfiguration narratives. Yet, the integration of local and global, individual and systemic perspectives is essential if one argues for their interdependence \cite{renn_evolution_2020}. We here discuss an approach to empirically tackle this challenge using methods developed within the framework of socio-epistemic networks (SEN) \cite{lerg_socio-epistemic_2024}. The framework is built on the assumption that knowledge is codified through the formation of cognitive, material, and social structures, representable via multi-layered, time dependent networks \cite{renn_netzwerke_2016}. Focusing on language-based changes, we here primarily address integration at the cognitive level, or in terms of SEN, the semantic layer. 

Our approach compares individual trajectories against the global development of scientific fields, examining how individual scholars’ works evolve alongside their disciplines. Adaptable to other domains and languages, we employ a data-driven methodology, applying (1) information-theoretic measures (relative entropy) on unigram language models and (2) estimating densities of document embeddings to analyse change over time, with words and sentences as the primary units of analysis. As an example of the global development of a scientific field, we look at the so-called \enquote{Renaissance of General Relativity} \cite{blum_renaissance_2016,will_renaissance_1989}, and use physicists Joseph Silk and Hans-Jürgen Treder as case studies of individual trajectories within this field. Although combining a quantitative view of structural changes with local case studies is necessarily incomplete, we argue that such methods facilitate fruitful comparison by highlighting tendencies or certain aspects of individual trajectories, such as language use within a larger development. 

We proceed with an overview of related work on the Renaissance of General Relativity and structural explanations (see \ref{subsec:Socio-epistemic Networks and the Renaissance of General Relativity}), corpus comparison, and diachronic language change (see \ref{subsec:Comparison of corpora and Diachronic Change}). Next, we introduce our data (see \ref{subsec:Data - NASA/ADS Corpus}) and methods (see \ref{subsec:Methods - Approaches for building Trajectories of Change}), followed by analyses of synchronous and asynchronous comparisons via relative entropy (see \ref{subsec:Method 1 - Changing Frequencies}) and density changes (see \ref{subsec:Method 2 - Changing Densities}). We conclude in  \ref{sec:Conclusion}. 

\section{Related Work}
\subsection{The Renaissance of General Relativity and Socio-epistemic Networks}\label{subsec:Socio-epistemic Networks and the Renaissance of General Relativity}
The \enquote{Renaissance of General Relativity} describes the resurgence of interest and research in the theory of general relativity, beginning in the mid 1950s and gaining momentum through the 1960s and 1970s, after a period of relative stagnation \cite{will_renaissance_1989}. During this \enquote{low-water mark} as \citet{eisenstaedt_low_1989} termed it, general relativity was largely seen as a niche area, providing minor corrections to Newtonian physics, with limited engagement from the broader physics community \cite{blum_reinvention_2015}. Several narratives have been proposed to explain this renaissance. One emphasizes the role of new empirical observations, particularly in astrophysics, such as the discovery of quasars or the cosmic microwave background \cite{will_renaissance_1989,blum_renaissance_2016}. Other narratives focus on the influx of funding and talent into physics after World War II, particularly in the United States, which lead to an immense growth in many areas of theoretical physics, including general relativity and gravitation research (GRG) \cite{kaiser_price_2018,lalli_building_2017}. A third narrative highlights the development of new theoretical tools that facilitated calculations and provided further insights into the theory, such as the Petrov classification or ADM formalism \cite{lalli_building_2017}.

A more integrated historiographical framework of the renaissance, proposed by \citet{blum_reinvention_2015}, argues for the interplay of both internal (epistemic) and external (social, institutional and material) factors in shaping the development of GRG as a scientific field in its own right. \citet{blum_reinvention_2015,blum_renaissance_2016,blum_gravitational_2018} show that the renaissance was not simply a reaction to external aspects but involved conceptual transformations within the theory itself, driven by the active formation of a research community. This community, initially dispersed geographically and across different research agendas (e.g., unified field theories, quantum gravity), coalesced around shared problems and slowly developed a common language and set of tools, enabling them to tackle fundamental questions about general relativity proper, such as the nature of gravitational waves or singularities. This historiographical framework was further developed and translated into a quantitative, network-oriented approach of so called socio-epistemic-networks (SEN) \cite{renn_evolution_2020,lerg_socio-epistemic_2024}.

Quantitative analyses using the SEN framework have provided further insights into causes of the Renaissance. \citet{lalli_dynamics_2020} employed network analysis to examine its social, semiotic/material, and semantic layers, revealing a significant increase in connectivity among researchers and institutions from the mid-1950s to the early 1960s, which preceded key astrophysical discoveries \cite{lerg_socio-epistemic_2024,blum_gravitational_2018}. The analysis again shows that this process was actively driven by community-building efforts, such as conferences and the establishment of international organisations, rather than solely a response to external stimuli. Analysis of co-citation networks further reveal a shift from attempts to replace general relativity with an encompassing theory to exploring the physical implications of the theory itself \cite{lalli_socio-epistemic_2020,blum_gravitational_2018}. This shift reflects the conceptual transformations stated by \citet{blum_reinvention_2015}, where GRG became a field in its own right. Methodologically, \citet{lalli_dynamics_2020,lalli_socio-epistemic_2020} used collaboration networks based on co-authorship, institutional affiliations, and/or conference participation to understand community formation, while co-citation networks were employed to analyse citation patterns in publications to identify research fronts, and trace the evolution of key topics by building semantic networks on top of co-citation clusters. 

One epistemic assumption of both the proposed historiographical and SEN frameworks is that analysing the evolution of knowledge systems is an analysis of the formation of structures. Both frameworks aim to link micro and macro-history by combining \enquote{the precision of micro-history grounded in ‘thick’ descriptions of historical contexts with the sweeping generalizations of global history, validating such generalizations through rich historical data} \cite[p. 319]{renn_evolution_2020}. However, they do not provide any operationalisation for implementing this integration. We here propose an approach using two compatible methods to achieve this by tracing micro or individual-history within the global formation of a research field, using language variation in comparing corpora.

\subsection{Comparison of Corpora and Diachronic Language Change}\label{subsec:Comparison of corpora and Diachronic Change}
Various domain-specific methods exist for comparing corpora in terms of linguistic and topic variation, structural analysis, or diachronic change. These include plagiarism detection, genre/register studies, semantic shift detection, corpus alignment for translation studies, stylometry and more. The work presented here focuses on computational approaches for diachronic comparison between individual and knowledge system corpora, situated at the intersection of corpus linguistics and quantitative corpus analysis. In these domains classical approaches often used probabilistic distributional representations of text using statistical methods, like frequency measures, to quantify linguistic variation \cite{kilgarriff_comparing_2001,juola_time_2003}, or to trace linguistic shifts in large-scale studies \cite{bochkarev_universals_2014}. In more recent studies distributional measures from information theory, such as Kullback-Leibler divergence (KLD) \cite{kullback_information_1951} and Jensen-Shannon divergence (JSD) \cite{lin_divergence_1991}, have also been employed to explore textual variation, detect stylistic changes, or to track linguistic shifts \cite{hughes_quantitative_2012,klingenstein_civilizing_2014,peter_fankhauser_exploring_2014}. But unlike traditional methods relying on predefined features \cite{van_kemenade_historical_2006,biber_styles_1989}, these measures offer more flexibility, i.e. KLD and JSD help to identify linguistic variation without pre-selecting periods, thereby potentially reducing the introduction of confirmation bias \cite{barron_individuals_2018,degaetano-ortlieb_using_2018}.

The first approach in this paper is build on them and is inspired by the project \enquote{Information Density in English Scientific Writing: A Diachronic Perspective} from the \href{https://sfb1102.uni-saarland.de/projects/information-density-in-english-scientific-writing/}{Collaborative Research Center (SFB 1102) Information Density and Linguistic Encoding (IDeaL)}, which uses entropy and surprisal\footnote{Surprisal, the basic concept underlying these measures, quantifies the predictability of linguistic units, measuring the information they carry \cite{eisenstein_bayesian_2008,degaetano-ortlieb_toward_2022}.} to investigate linguistic change. Analysing the Royal Society Corpus (RSC) \cite{fischer_royal_2020,kermes_royal_2016}, \citet{bizzoni_linguistic_2020,degaetano-ortlieb_using_2018,degaetano-ortlieb_information-theoretic_2019,degaetano-ortlieb_measuring_2021,degaetano-ortlieb_toward_2022,teich_less_2021} quantitatively show that scientific writing from the 17th to 19th centuries evolved toward denser information encoding and greater standardisation, as previously qualitatively argued by \citet{halliday_writing_1993}. They observe a dynamic interplay between diversification (lexical innovation) and conventionalisation (the emergence of predictable grammatical structures), resulting in the development of an \enquote{optimal code} for expert communication \cite{degaetano-ortlieb_measuring_2021,degaetano-ortlieb_toward_2022}. Similar effects have been identified in cognitive science and psycholinguistics \cite{delogu_teasing_2017,levy_expectation-based_2008}. More than being able to identify significant linguistic changes, KLD highlights the specific features driving these shifts \cite{bizzoni_linguistic_2020,degaetano-ortlieb_using_2018}. Since we are interested in continuos, period-independent changes over time as well as the features driving these changes, we will adapt this flexible, data-driven approach here for our purposes.

Another classical approach for quantitative corpus analysis are probabilistic topic models \cite{steyvers_probabilistic_2014}, widely used to uncover thematic structures and shifts \cite{blei_dynamic_2006,griffiths_finding_2004,hall_studying_2008,mcfarland_differentiating_2013,yang_topic_2011}. These models successfully identify topical diversification in diachronic corpora but are limited by their bag-of-words assumption, ignoring order, syntax, and context. Neural-based approaches, like word embeddings, have advanced corpus comparison, particularly in studying lexical semantic change. However, early static models like Word2Vec or GloVe learn word vectors based on co-occurrence, i.e. still from probabilistic distributional patterns \cite{mikolov_efficient_2013,pennington_glove_2014}. While useful for tracking semantic shifts \cite{kulkarni_statistically_2015,dubossarsky_outta_2017}, static embeddings assign fixed vectors to words, failing to capture polysemy or contextual variation. Contextualised embeddings from models like ELMo, BERT, and RoBERTa \cite{devlin_bert_2019,liu_roberta_2019,peters_deep_2018} generate dynamic word representations based on context, therefore capturing polysemy and semantic change at a more granular level. Sentence-BERT \cite{reimers_sentence-bert_2019} or GPT \cite{radford_improving_2018} Models extend this to sentence and document embeddings. Consequently, contextualized embeddings are used in most of the recent studies on semantic shift detection \cite{zichert_tracing_2024,Montanelli2023,giulianelli_analysing_2020,conneau_unsupervised_2020}. They have also been integrated into topic modelling pipelines, e.g., BERTopic \cite{Grootendorst2022}, combining topic modelling with the mentioned contextual advantages of these models. The rapid growth of domain-specific embedding models like SciBERT \cite{beltagy_scibert_2019} or PhysBERT \cite{hellert_physbert_2024} further enhance genre specific capabilities to capture semantics in specialised literature. Our second approach builds on these advances by using document embeddings, offering an alternative but compatible route to the first approach, based on transformer-based, rather than probabilistic, distributional representations.

In summary, we address the challenge of tracking individual versus system knowledge evolution by (a) adapting and extending established computational linguistic methods, (b) combining them with context-sensitive transformer-based approaches for greater robustness, and (c) comparing these methods for two individuals within the field of GRG to evaluate the consistency of both methods.

\section{Data and Methods}\label{sec:Data and Methods}

\subsection{Data - NASA/ADS Corpus}\label{subsec:Data - NASA/ADS Corpus}
For the case discussed here, a corpus of publications related to the field of General Relativity and Gravitation (GRG) from 1911 to 2000 is used, whereby our selection criteria are based on those of the project \enquote{\href{https://www.mpiwg-berlin.mpg.de/project/renaissance-general-relativity}{The Renaissance of General Relativity in the Post-World War II Period}} at the Max Planck Institute for the History of Science \cite{blum_renaissance_2020}. This alignment allows for the integration and comparison of our findings with those of the project (see subsection \ref{subsec:Socio-epistemic Networks and the Renaissance of General Relativity}). To identify authors who wrote in the period 1925-1970 on topics related to general relativity, \citet{lalli_dynamics_2020} created a set of articles and actors primarily based on data from the Web of Science, which led to 8,296 articles in their GRG publication space. In contrast, we employ the NASA/ADS Astrophysics Data System, which offers a more comprehensive coverage of GRG, particularly for non-English publications. Still, the dataset shares similar problems most bibliographic datasets built from online repositories have, i.e. incompleteness (e.g., full text, citation data), bias (e.g., language/geographic, collection focus), and translation inaccuracies. Additionally, our corpus is not balanced, as later years contain substantially more publications. This is due not only to the increasing volume of scientific output but also to collection effects, such as the mentioned language and geographic biases, which are less pronounced in more recent publications due to the growing dominance of English as the language of science and the homogenization and globalization of publication and indexing practices. All these factors call for caution when interpreting frequency and similarity effects. We extended the time frame to 1911-2000, using comparable search criteria\footnote{We combined article sets from various queries to create a comprehensive dataset representing the general relativity publication space from 1911 to 2000, akin to \citet{lalli_dynamics_2020}. \textbf{Set A} consists of hand-curated works by Albert Einstein (1911-1955) related to general relativity, cosmology, and unified field theory. \textbf{Set B} includes all articles citing Set A in the period 1915-2000. \textbf{Set C} includes all articles citing Set B (1915-2000). Together, these three sets form \textbf{Set D} (Einstein’s citation space). We extended the dataset using keyword searches to form \textbf{Set E}, which includes works related to GRG (1915-2000). \textbf{Set F} combines Set D and Set E, representing the general relativity publication space.}. 1911 is chosen, as it marks Einstein's first publications on the theory that would become general relativity in 1915. Our dataset contains around 180,000 publications in total; however, for the comparative analysis of the chosen cases in this paper (see \ref{subsec:Case Study}), we focus only on publications from 1957 onwards, as this is the year of Treder's first listed publication (Silk's is 1962). For both methods, we analyse the textual content of our documents, which in this case consist of titles and abstracts translated into English. For Method 1, we use the concatenated, translated texts, with numbers, symbols, and stop words removed, followed by tokenization and lemmatization. For Method 2, we use the concatenated, translated texts without these additional steps. The corpus is made available under a Creative Commons license\footnote{\href{https://doi.org/10.5281/zenodo.14581502}{https://doi.org/10.5281/zenodo.14581502}}.

\subsection{Methods - Approaches for building Trajectories of Change}\label{subsec:Methods - Approaches for building Trajectories of Change} 
To examine the diachronic development of individual vs. global for this corpus and specifically analyse (language-based) changes on the semantic layer, we employ two different but complementary methods. The first one uses relative entropy or Kullback-Leiber divergence (KLD) and the second applies sentence Embedding Density Estimations (EDE). By doing so, we will investigate whether the tendencies of the results are consistent across both methods or if they differ, and if so, in what aspects. The aim of this multi-method approach is to increase the robustness of the analysis by offsetting the weaknesses of one approach with the strengths of another, which ideally increases the chances of a more accurate historical analysis. 

\subsubsection{Method 1 - Information theoretical approach}\label{subsubsec:Method 1 - Information theoretical approach}
KLD measures the difference between two probability distributions. In our context of text analysis, it quantifies how one text (e.g. parts of a corpus) diverges from another based on term distributions, where a lower KLD indicates greater similarity between the two. The KLD from model \( M_q \) to model \( M_d \) is defined as:
\begin{equation}
D\left(M_d \| M_q\right)=\sum_{t \in V} P\left(t \mid M_d\right) \log _2 \frac{P\left(t \mid M_d\right)}{P\left(t \mid M_q\right)}
\end{equation}
and calculates the average number of additional bits needed to encode terms from \( M_d \) using the probability distribution of \( M_q \). Since it is a sum, it effectively captures which terms contribute most to the divergence, also allowing for the identification of terms that are particularly characteristic or \enquote{typical} of one corpus over another \cite{degaetano-ortlieb_information-based_2016}. \( M_d \) and \( M_q \) here are unigram language models, which assign probabilities to individual terms based on their relative frequencies in a text sequence, assuming each term occurs independently of others. This means the probability of a text sequence is calculated as the product of the probabilities of its individual words, estimated from their relative frequencies within the text. Texts are thus represented as probability distributions. A problem this approach encounters occurs when certain words present in one corpus are absent in the other, leading to zero probabilities that can heavily distort KLD calculations. This issue of out-of-vocabulary (OOV) words is addressed, by applying smoothing techniques, here Jelinek-Mercer smoothing (with lambda 0.05), which adjusts probability estimates to assign small, non-zero probabilities to unseen words, thereby preventing zero probabilities in the overall model\footnote{Jelinek-Mercer smoothing interpolates between the observed probabilities from one text sequence and the overall probabilities in the entire corpus. It assigns a small probability to unseen words, reducing the impact of zero probabilities.}.

We use the \enquote{general relativity publication space} as a representation of the field of General Relativity and Gravitation (GRG) and the publications of the individuals or groups under study, which are part of this publication space, to represent their work within the same timeframe. We divide the considered period 1957-2000 into intervals of two years, and generate the models for both the field and the individuals for each time slice. We then calculate the pointwise and summed KLD between these models synchronously for each interval, allowing us to plot the divergence per slice over time and respectively identify highs and lows that may indicate significant deviations (see \ref{subsubsec:Synchronous Comparison - Summed}). Before the summation, we apply an unpaired Welch's t-test (p-value < 0.05) to each term to evaluate whether the differences in relative term frequencies between the two corpora are statistically significant, accounting for variations in sample sizes and variances. We sort the pointwise KLD values and filter the terms that pass the significance test, allowing us to pinpoint the specific terms that most differentiate the individual's work from the broader field at various time slices (see \ref{subsubsec:Synchronous Comparison - Point-wise}). Additionally, we perform asynchronous comparisons by calculating the KLD between each time slice and all other slices, which helps us identify periods where the divergence is minimal or maximal (see \ref{subsubsec:Asynchronous Comparison - Summed}). For instance, if the divergence between a researcher's publications in 1970 and the field's publications in 1946 is minimal, it suggest a higher lexical similarity between these two periods than between any other\footnote{The described methods are available as part of the \href{https://semanticlayertools.readthedocs.io}{semanticlayertool} package, developed by the \href{https://modelsen.gea.mpg.de/}{ModelSEN} project. They can be accessed via the classes \href{https://semanticlayertools.readthedocs.io/en/latest/linkage.html\#semanticlayertools.linkage.worddistributions.UnigramKLD}{UnigramKLD} and \href{https://semanticlayertools.readthedocs.io/en/latest/linkage.html\#semanticlayertools.linkage.densities.EmbeddingDensities}{EmbeddingDensities}}.

\subsubsection{Method 2 - Density approach}\label{subsubsec:Method 2 - Density approach}
Secondly, we look at local versus global variations in document embedding densities. As mentioned in \ref{subsec:Comparison of corpora and Diachronic Change}, document embeddings are numerical vector representations of texts generated by language models like BERT or GPT. They capture a document's semantic and syntactic content in a fixed-length, high-dimensional vector, allowing for comparisons based on meaning rather than word frequencies, enabling very effective clustering or classification. Similar to BERTopic \cite{Grootendorst2022}, we convert our corpus into embedding vectors and reduce their dimensions for better visualization. We then identify clusters and label them using representative documents, resulting in a time-dependent visualization where similar publications are grouped closely. After experimenting with several models, including SciBERT \cite{beltagy_scibert_2019}, PhysBERT \cite{hellert_physbert_2024}, and OpenAI’s \texttt{text-embedding-3-large}, we found \texttt{text-embedding-3-large} generated embeddings that were most accurate and meaningful in terms of representation tuning.

To track changes over time, we employ Kernel Density Estimation (KDE) to analyse density variations around each embedding vector. KDE is a non-parametric method for estimating the probability density function of a random variable based on a finite sample. It provides a smooth estimate of the data's distribution, allowing us to assess how densely documents are clustered in different regions of embedding space.

For each of the individual's publications, we follow these steps:

\begin{enumerate}
    \item  We extract the embedding vector of the publication as the reference document for density estimation.
    \item Starting from the publication year, we calculate KDE in two-year increments. For each time slice:, we:
    \begin{enumerate}
        \item Collect embeddings of all documents published within that time slice.
    \end{enumerate}
    \begin{enumerate}
        \item  Include the reference document in each time slice's dataset, even if not published in that period, to measure how the field evolves relative to this fixed reference. 
    \end{enumerate}
    \begin{enumerate}
        \item  Fit a KDE model to the embeddings using a Gaussian kernel.
    \end{enumerate}
    \item For each period, we estimate the density at the reference document using the fitted KDE model. The density function is:
\begin{equation}
    \hat{f}(x) = \frac{1}{n h} \sum_{i=1}^{n} \frac{1}{\sqrt{2\pi}} \exp\left( -\frac{1}{2} \left( \frac{x - x_i}{h} \right)^2 \right)
\end{equation}
 where \( \hat{f}(x) \) is the estimated density at reference document \( x \), \( n \) is the number of publications in the time slice, \( h \) controls smoothness, and \( x_i \) are the embedding vectors of other documents.  
    \item We record the estimated densities over time, which allows us to observe how the \enquote{neighbourhood} around the document evolves—whether the concentration of similar content increases, remains steady, or decreases. 
\end{enumerate}

By including the individual's publication in each time slice's KDE model, we track how the field’s focus shifts relative to the topic of that publication. Increasing Embedding Density Estimation (EDE) at the reference document suggests that documents with similar content are being published later, indicating the topic's growing relevance. Conversely, decreasing EDE suggests the topic is becoming less central. This process helps us monitor the thematic centrality of an individual's publications over time and how they align with or diverge from broader trends. We also calculate the median EDE for each time slice across all of the individual's publications, offering an aggregate view of their topical centrality. A rising median EDE suggests the individual’s topics are becoming more central to ongoing research.

This approach complements our earlier use of Kullback-Leibler Divergence by offering a different perspective on topic evolution. While KLD measures divergence in term distributions between corpora, this method captures semantic similarity at the document level, accounting for contextualized term usage. It allows us to trace trajectories of individual or groups of documents, focusing on either content (topic combinations) or metadata (documents from specific authors, institutions, geographic locations, etc.).

\section{Analysis}\label{sec:Analysis}
The analysis is based on the assumption that if an individual’s terminology closely aligns with that of the mainstream (low KLD), many researchers are likely addressing similar topics to those the individual is addressing (high EDE around individual publications), and vice versa. We will apply the two methods described (\ref{subsubsec:Method 1 - Information theoretical approach} and \ref{subsubsec:Method 2 - Density approach}) to the GRG corpus (\ref{subsec:Data - NASA/ADS Corpus}) to test this assumption. We begin by introducing the two individuals selected for this analysis.

\subsection{Case Study}\label{subsec:Case Study}
The cases we test these two methods on are drawn from the history of General Relativity and Gravitation (GRG) research. As mentioned in \ref{subsec:Socio-epistemic Networks and the Renaissance of General Relativity}, GRG as a field began to form and gain traction only after World War II. This \enquote{Renaissance of General Relativity}, overlaps with the most productive years of our two individual cases. Furthermore we choose these two authors, since they are comparable in our dataset, in terms of publications size and time-frame, but represent different research cultures and geopolitical spheres. 

Our first case is Joseph Silk (born 1942), a British-American astrophysicist. Educated at the University of Cambridge and earning his Ph.D. from Harvard University in 1968, Silk held positions at the University of California, Berkeley, and later at the University of Oxford as the Savilian Professor of Astronomy. As of 2024, he is a professor at the Institut d'Astrophysique de Paris and Homewood Professor of Physics and Astronomy at Johns Hopkins University. Silk is best known for introducing \enquote{Silk damping}, a concept describing how photon diffusion in the early universe smoothed out density fluctuations, which impacted understanding of the cosmic microwave background (CMB) and marked cosmology’s transition into a high-precision science. His extensive research on dark matter, galaxy formation, and mass loss dynamics influenced theoretical astrophysics but also advanced new observational technologies. Embracing empirical and theoretical integration, Silk's work aligns with the global shift toward data-driven, international, collaborative research facilitated by advancements in observational technology. He is one of the most cited authors in our dataset, as well as one of the authors with the most publications listed in it.  

Hans-Jürgen Treder (1928–2006), a German physicist, by some referred to as the \enquote{Einstein of the GDR}, exemplifies a more principle-oriented approach to GRG. He studied mathematics, physics, and philosophy in Berlin, earning his doctorate in 1956 and became Professor at Humboldt University and Director of the Institute for Pure Mathematics in 1963. Later, he directed the Central Institute for Astrophysics and the Einstein-Laboratorium in Potsdam. Treder made contributions to general relativity, cosmology, and efforts to unify micro- and macrophysics. His lifelong dedication to foundational questions in physics, where he sought to bridge physical principles with philosophical and epistemological ones was heavily influenced by Einstein’s ideas, and the so called \enquote{Große Berliner Physik} tradition of Helmholtz, Planck, and others \cite{hoffmann_prinzipien_2022}. Among Treder's most known contributions are therefore principle-based approaches, like the \enquote{Bezugstetraden-Theorie}, a tetrad-based framework allowing alternative gravity-matter couplings to address gravitational collapse, and his work on \enquote{inertia-free mechanics} and the Mach-Einstein doctrine, which sought to connect gravity with inertia. His intellectual legacy is complex; despite his ambitious ideas, most of Treder’s work remained peripheral internationally, partly due to dynamics of the Cold War and his adherence to principle-based physics over model-driven approaches \cite{hoffmann_prinzipien_2022,schlattmann_relativity_2020}. He is one of the most frequently represented authors in our corpus as well. 

\subsection{General statistics for both authors}\label{subsec:General statistics for both authors}
Treder and Silk have comparable publication counts in our dataset, with Treder contributing 355 publications and Silk 332. However, there is a notable difference in citation metrics: Silk’s publications show a much higher average citation count (51.4 vs. Treder’s 2.08), with cumulative citations of 17,064 for Silk and 738 for Treder. This disparity likely reflects a combination of factors. According to our ADS data, Silk’s work often appeared in journals like \textit{Nature} and \textit{The Astrophysical Journal}, whose international distribution and influence grew over time. Treder primarily published in \textit{Annalen der Physik} and \textit{Astronomische Nachrichten}, which historically held significant influence but did not grow comparably\footnote{Data is taken from Exaly, an open-access scientometric platform providing citation analysis and impact metrics. As mentioned above, it also faces challenges such as distinguishing citable documents and handling non-English text, which may result in underestimation biases. See Impact Factor for: \href{https://exaly.com/journal/12348/nature/impact-factor}{Nature}, \href{https://exaly.com/journal/13854/astrophysical-journal/impact-factor}{The Astrophysical Journal}, \href{https://exaly.com/journal/12627/annalen-der-physik/impact-factor}{Annalen der Physik}, \href{https://exaly.com/journal/12547/astronomische-nachrichten/impact-factor}{Astronomische Nachrichten}}, presumably partly due to their mixed language (English, non-English) content (see Table \ref{tab:top_journals}). 

\begin{table}[ht!]
\caption{Top Journals by Publication Count for Treder and Silk.}
\vspace{0.5cm}
\centering
\begin{tabular}{lcc}
\hline
\textbf{Journals} & \textbf{Treder (Count)} & \textbf{Silk (Count)} \\
\hline
\textit{Annalen der Physik} & 98 & - \\
\textit{Astronomische Nachrichten} & 47 & - \\
\textit{Foundations of Physics} & 33 & - \\
\textit{The Astrophysical Journal} & - & 125 \\
\textit{Nature} & - & 30 \\
\hline
\end{tabular}
\label{tab:top_journals}
\end{table}

Concurrently, Treder’s use of German (173 publications in English, 175 in German) may have limited his international reach as English became the dominant language of global science in this period. Further factors may include Cold War constraints and East German science policy, which restricted access to increasing international collaborations, such as the so-called \enquote{postdoc cascade}, where international exchanges during postdoctoral training facilitated the dissemination of local knowledge and increase of international collaboration \cite[60 ff.]{kaiser_drawing_2005}. This contrast is also evident in co-authorship numbers: Treder’s most frequent collaborators are affiliated with German institutions, suggesting a more regionally focused network, while Silk’s collaborators are more internationally dispersed, suggesting a higher integration into a global scientific community. 

\begin{table}[ht!]
\caption{Top nationalities of co-author affiliations for Treder and Silk.}
\vspace{0.5cm}
\centering
\begin{tabular}{lcc}
\hline
\textbf{Country} & \textbf{Treder (Count)} & \textbf{Silk (Count)} \\
\hline
Germany & 13 & 3 \\
United States & - & 21 \\
United Kingdom & - & 6 \\
France & - & 5 \\
Canada & - & 2 \\
Netherlands & - & 2 \\
\hline
\end{tabular}
\label{tab:top_nationalities}
\end{table}

These structural factors may be amplified by limited metadata indexing of non-English-language journals in major citation databases, potentially distorting the dataset’s reference and citation metrics. For example, a manual review of Treder's publications within a reference corpus of his complete works suggests that references to Treder's work are significantly under-reported in our dataset (1.83 per publication compared to 16.69 in manually verified data), likely due to incomplete indexing of German-language journals.

\subsubsection{Relative Token Usage}
\begin{figure}[ht!]
    \centering
    \begin{subfigure}[b]{0.5\textwidth}
        \includegraphics[width=\linewidth]{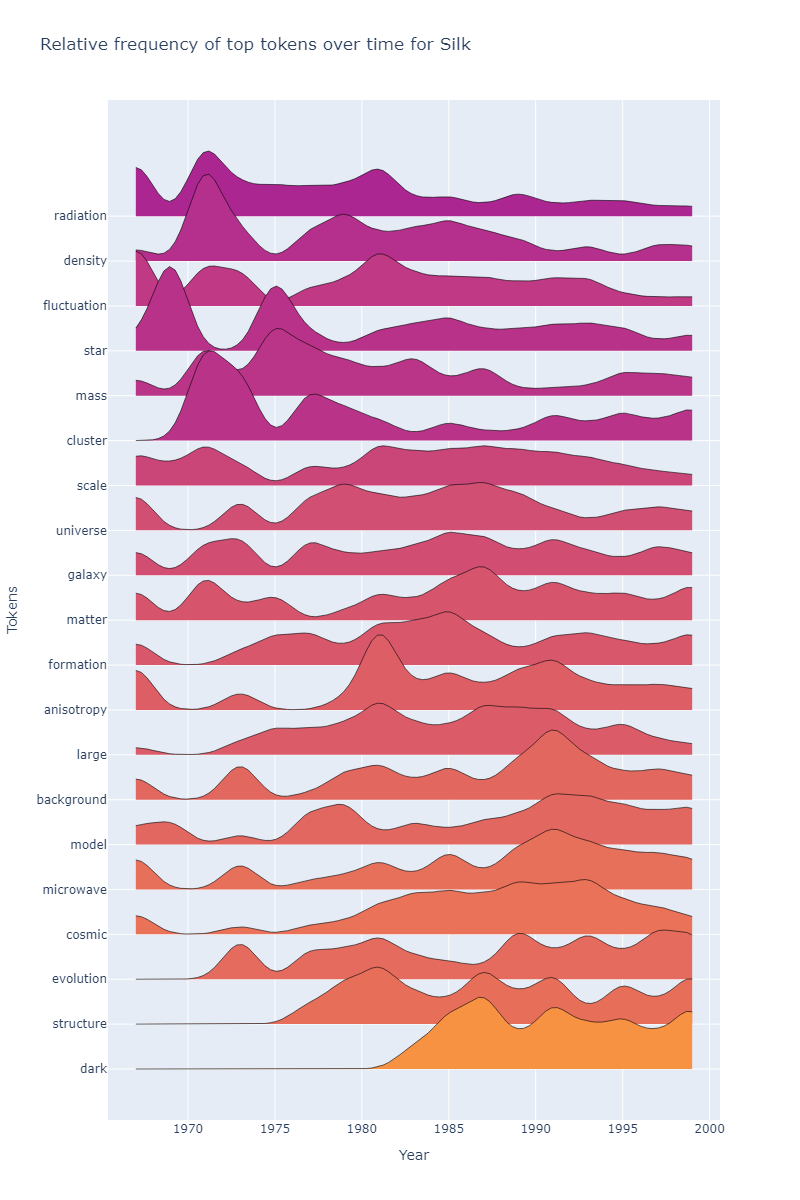}
    \end{subfigure}%
    \begin{subfigure}[b]{0.5\textwidth}
        \includegraphics[width=\linewidth]{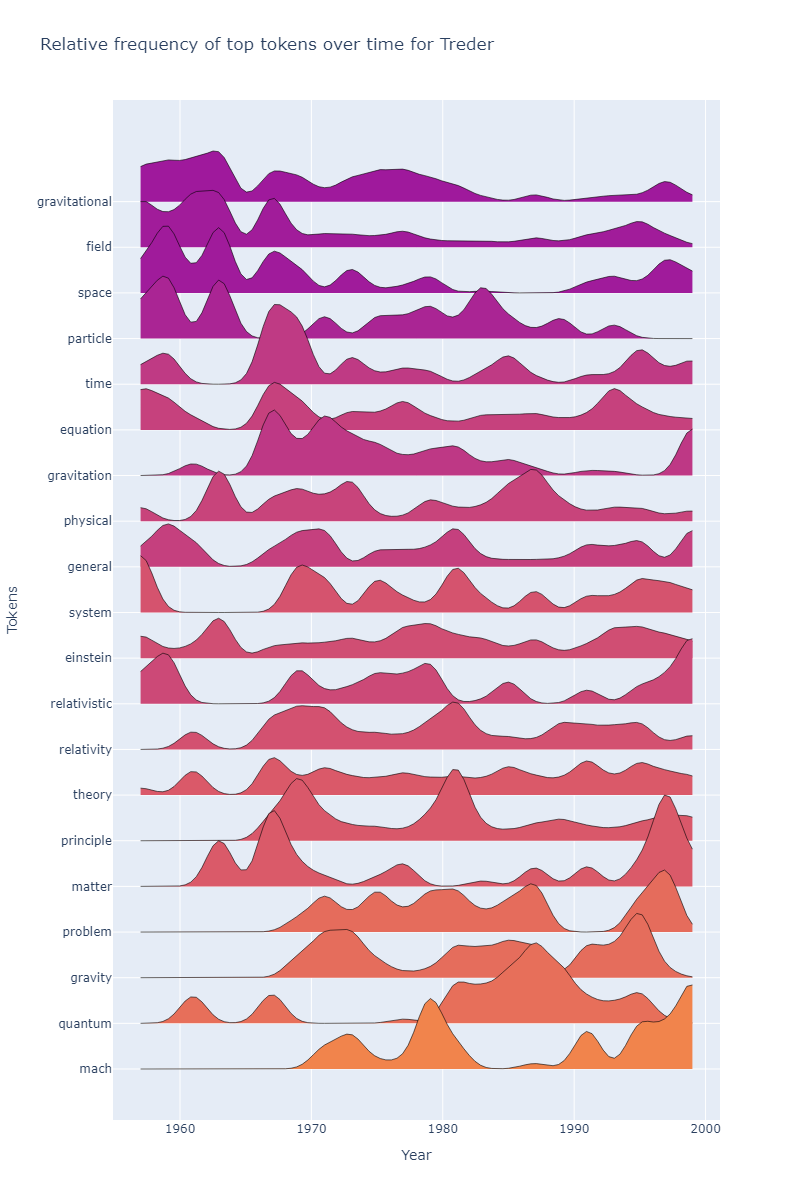}
    \end{subfigure}
    \caption{Relative Token Usage over Time for Silk (left) and Treder (right). The relative frequency of each token is calculated by dividing its occurrence by the total number of tokens within each two-year interval, focusing on the 20 most frequently used terms. To reduce the impact of years with insufficient data, any intervals with fewer than 50 total tokens were excluded. Terms were expanded, grouped by two-year bins, and scaled, with KDE applied to smooth frequencies for clearer trends.}
    \label{fig:relative_tokens}
\end{figure}

Figure \ref{fig:relative_tokens}, illustrates the trends of token usage over time for Silk and Treder, respectively. 

Silk’s token usage shows notable peaks in terms like \enquote{radiation}, \enquote{density}, and \enquote{star} during the late 1960s and early 1970s, reflecting an early focus on cosmic phenomena and matter density. In the 1980s, terms such as \enquote{cluster}, \enquote{microwave}, and \enquote{anisotropy} gain prominence, aligning with Silk’s engagement in studies of cosmic structures and the cosmic microwave background (CMB). By the 1990s, an increase in tokens like \enquote{dark} and \enquote{structure} points to Silk’s growing interest in dark matter and large-scale cosmic formation, mirroring shifts in cosmological research. The usage of the term \enquote{model} indicates an intensified focus on modelling.

Treder’s relative token usage, on the other hand, displays more consistent activity. Tokens such as \enquote{einstein}, \enquote{theory}, \enquote{general}, \enquote{relativity}, and \enquote{principle} maintain an almost steady presence from the 1960s onward, indicating Treder’s sustained focus on foundational issues in theoretical physics, particularly general relativity. Similarly, early peaks in terms like \enquote{particle}, \enquote{field}, and later \enquote{quantum} suggest his continued interest in connecting micro with macrophysics, albeit with shifts over time. The term \enquote{mach} shows distinct peaks once during the 1970s and again starting from the 1990s, hinting at his recurring focus on Mach's Principle and the so-called \enquote{Mach-Einstein Doctrine}.

Based on our data, Silk’s and Treder’s academic profiles reflect distinct thematic focuses, publication venues, language choices, and collaboration patterns. Silk’s alignment with international high-impact journals and collaborations as well as citations, gives a hint of his role in astrophysics and cosmology, while Treder’s more regional collaborations and principle-based focus point towards a consistent engagement with foundational topics in GRG, though seemingly with less reception. These observations, however, are to be understood as tendencies, shaped by biases in indexing, language, and citation practices.

\subsection{Method 1 - Changing Frequencies}\label{subsec:Method 1 - Changing Frequencies}
We now turn to our first method, outlined in \ref{subsubsec:Method 1 - Information theoretical approach}. First, we transform all publications from the respective individual in each time slice, along with the rest of the publications from that slice, representing the field, into probability distributions of lemma unigrams. Second, we calculate the pointwise and summed KLD between these models synchronously for each interval, enabling us to plot divergence per slice over time and identify significant peaks and troughs that may indicate deviations. Third, we calculate the summed KLD between these models asynchronously for all time slices, hinting at slices, where terminology used by the individual is most similar to that of the rest of the field.   

\subsubsection{Synchronous Comparison - Summed}\label{subsubsec:Synchronous Comparison - Summed}

\begin{figure}[ht!]
    \centering
    \begin{subfigure}[b]{\textwidth}
        \includegraphics[width=\linewidth]{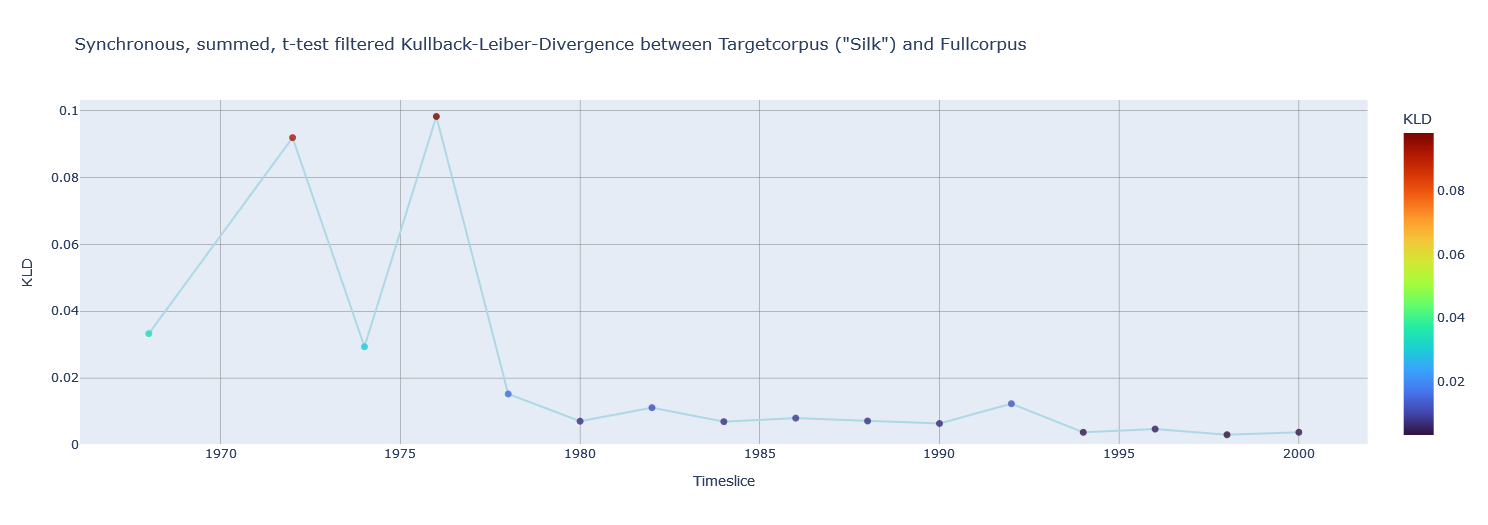}
    \end{subfigure}
    \begin{subfigure}[b]{\textwidth}
        \includegraphics[width=\linewidth]{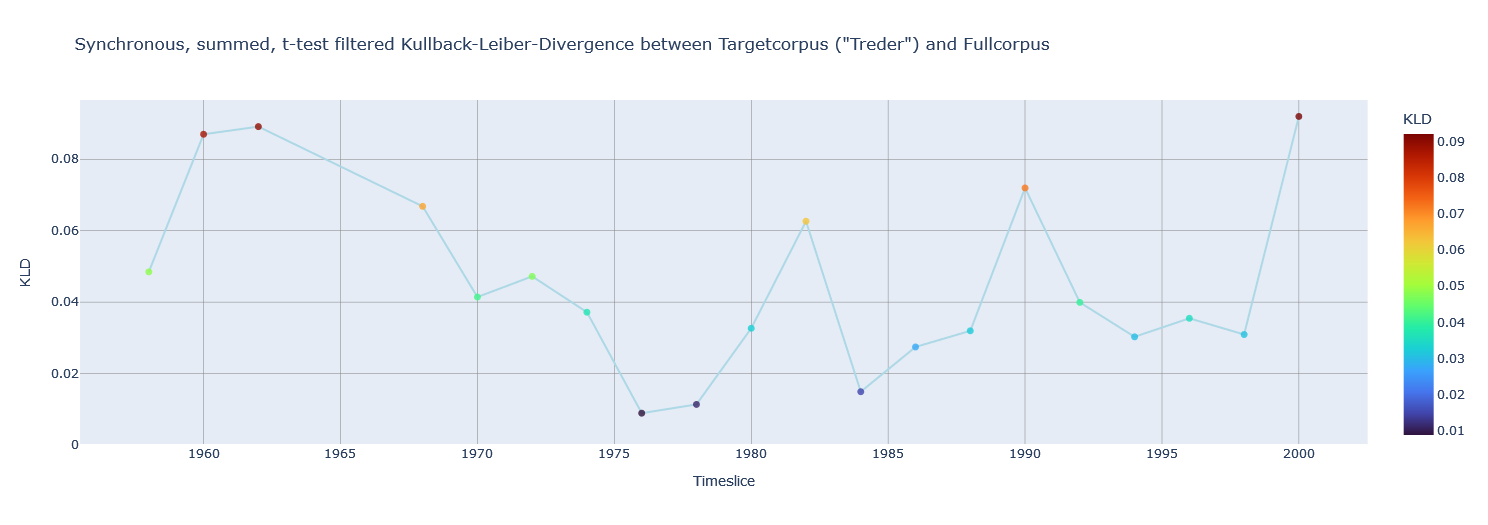}
    \end{subfigure}
    \caption{Summed, synchronous Kullback-Leibler Divergence (KLD) over time for Silk (top) and Treder (bottom), showing the development of divergence in summed term usage relative to the full corpus. Using non-overlapping time slices, unigram models were generated for each slice, applying Jelinek-Mercer smoothing ($\lambda = 0.05$) to avoid zero probabilities, retaining only high significance terms, filtered via Welch's t-test ($\alpha = 0.05$).}
    \label{fig:synchronous_summed_kld}
\end{figure}

Figure \ref{fig:synchronous_summed_kld} compares Silk and Treder with the overall field. During the 1960s to mid-1970s, as disparate research agendas merged into a more coherent field during the Renaissance, inconsistent lexical usage transitions towards a shared or \enquote{mainstream} terminology.\footnote{Here, \enquote{mainstream} refers to all tokens from all publications in a time slice excluding those by the author(s) being compared; high-frequency tokens dominate this distribution, forming the mainstream.} At this stage, both Silk and Treder were relatively young researchers, embedded in their local research traditions, as seen in relatively high and, for Silk, fluctuating KLD values.

In the mid to late 1970s, both authors experience growth in productivity and prominence, with Treder entering his most influential period and Silk starting to produce approximately 15 papers per year. During this time, both exhibit low KLD values, indicating alignment with mainstream GRG terminology. However, their trajectories diverge significantly afterward: Silk, becoming one of the corpus’s most cited authors, maintains low KLD values, suggesting consistent use of mainstream terminology. In contrast, Treder increasingly diverges from this trend, reaching peak KLD around 2000. This divergence is also seen in a qualitative analysis , as briefly mentioned in \ref{subsec:Case Study}, where Treder's work, grounded in fundamental principles and frequently referencing philosophy and history, gradually shifted toward a niche within GRG research \cite{hoffmann_prinzipien_2022}.

\subsubsection{Synchronous Comparison - Point-wise}\label{subsubsec:Synchronous Comparison - Point-wise} 

\begin{figure}[ht!]
    \centering
    \begin{subfigure}[b]{\textwidth}
        \includegraphics[width=\linewidth]{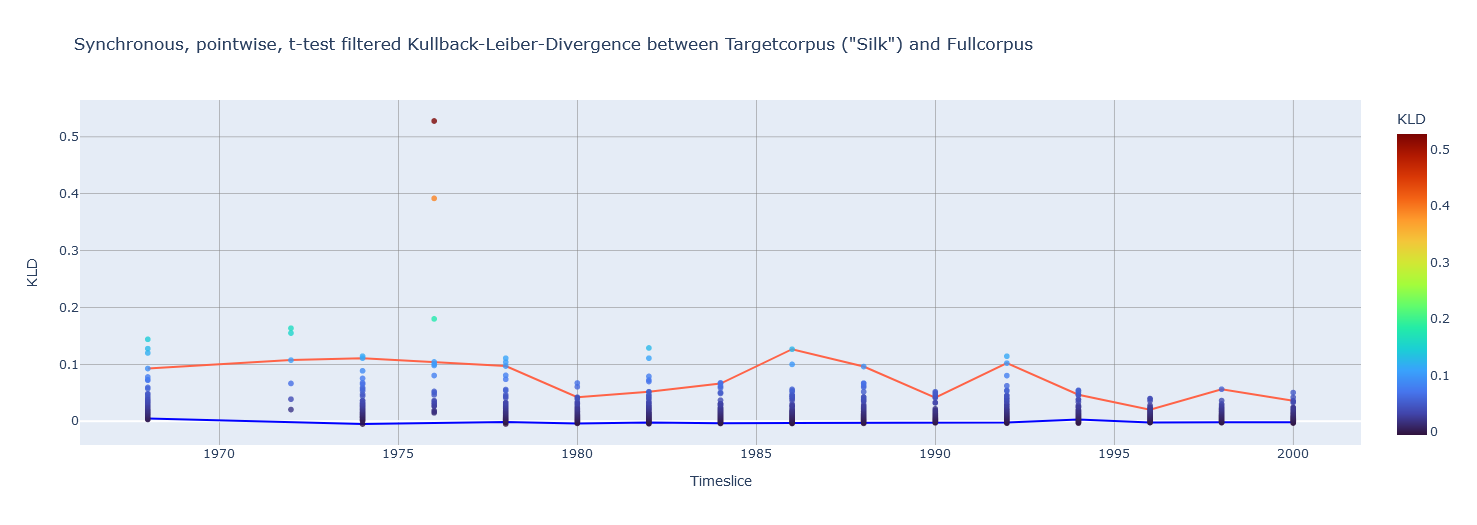}
    \end{subfigure}
    \begin{subfigure}[b]{\textwidth}
        \includegraphics[width=\linewidth]{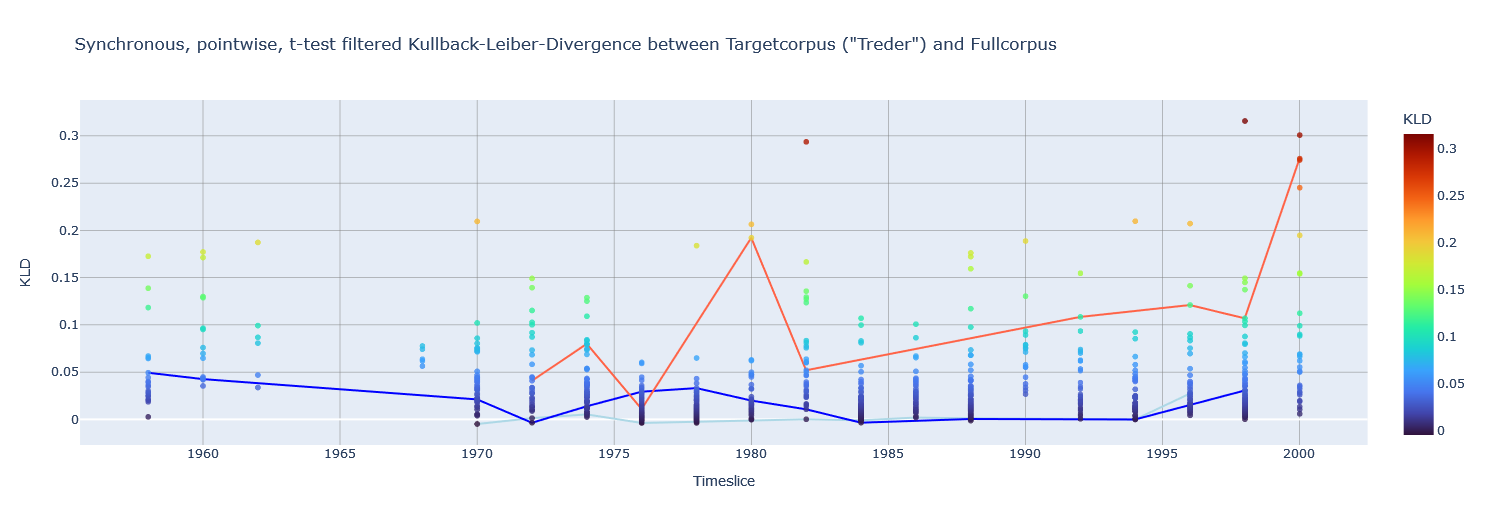}
    \end{subfigure}
    \caption{Synchronous, pointwise Kullback-Leibler Divergence (KLD) over time for Silk (top) and Treder (bottom), showing the development of divergence in individual term usage relative to the full corpus. The blue and light blue baselines represent terms with one of the lowest cumulative KLD values across all slices (\enquote{gravitational} and \enquote{gravity}). The red lines highlight terms with one of the highest cumulative KLD values across all slices (\enquote{galaxy} for Silk and \enquote{mach} for Treder).}
    \label{fig:synchronous_pointwise_kld}
\end{figure}

This divergence between the two authors is further evident when examining the specific words driving high KLD values. Figure \ref{fig:synchronous_pointwise_kld} shows point/word-wise divergence values over the same years. 

The blue and light blue baselines represent the terms \enquote{gravitational} and \enquote{gravity}, which maintain low divergence values throughout for both authors, as expected given the frequent mention of these terms in a corpus on gravitational physics. \enquote{Einstein} might also be expected to have a high frequency in a GRG corpus, yet Einstein’s theories and concepts were so ingrained that explicit mentions are actually less frequent overall. Treder, however, as mentioned in \ref{subsec:Case Study}, constantly references Einstein in his work, resulting in the highest summed divergence score for this term, followed by \enquote{principle}. The red line, indicating the word \enquote{mach}, also peaks at similar times as in our analysis of term usage (see \ref{subsec:General statistics for both authors}), indicating that engagement with topics around Mach's Principle and related ideas was specific to Treder's research rather than a trending topic in the broader community at that time. Silk’s overall highest-scoring KLD term is \enquote{galaxy}, again marked by the red line, showing his consistent use of it beyond mainstream levels, as might be expected of a specialist in astrophysics/cosmology.

\subsubsection{Asynchronous Comparison - Summed}\label{subsubsec:Asynchronous Comparison - Summed}

\begin{figure}[ht!]
    \centering
    \begin{subfigure}[b]{\textwidth}
        \includegraphics[width=\linewidth]{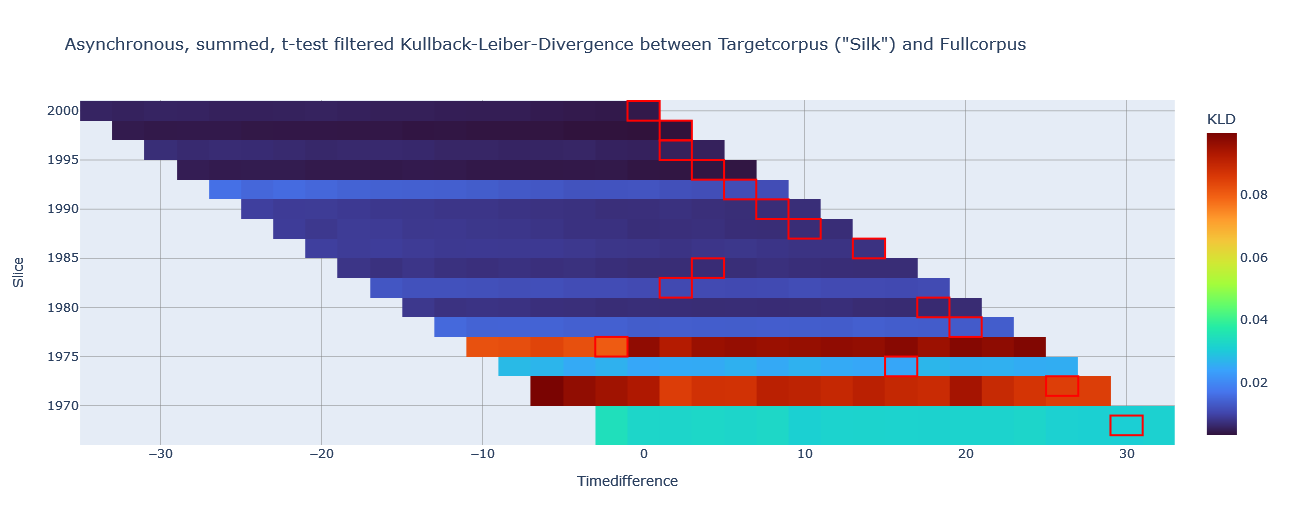}
    \end{subfigure}
    \begin{subfigure}[b]{\textwidth}
        \includegraphics[width=\linewidth]{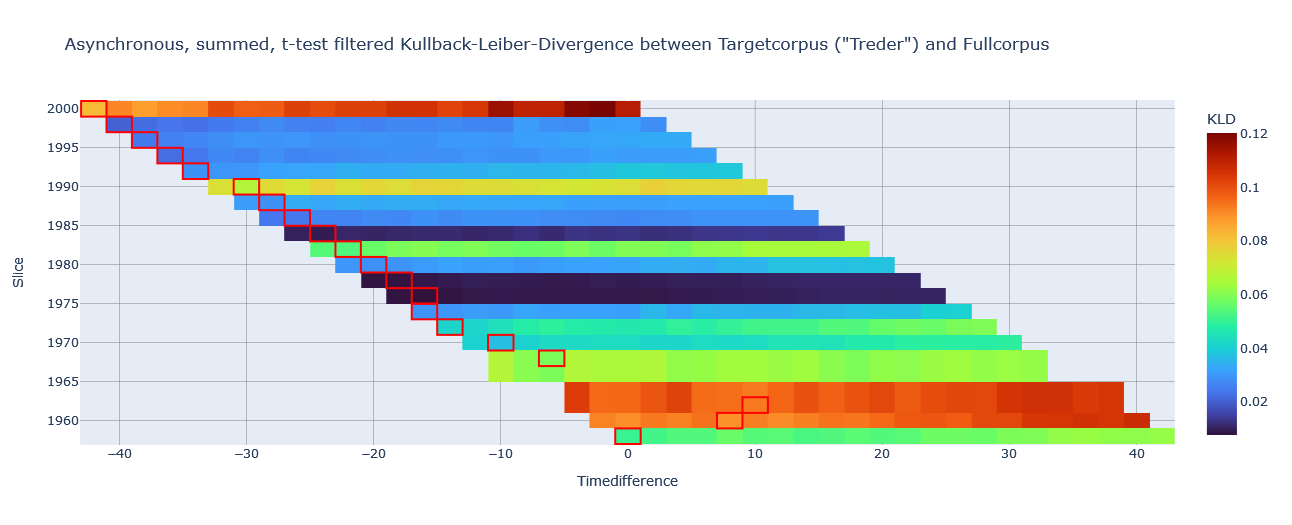}
    \end{subfigure}
    \caption{Asynchronous, summed Kullback-Leibler Divergence (KLD) for Silk (top) and Treder (bottom), showing divergence in term usage for each time slice relative to all other slices in the full corpus. On the x-axis is the time difference, showing comparisons of each time slice in the individual corpus to all others in the full corpus. The y-axis represents the time slices of the individual corpus. Years with the lowest divergence values are highlighted in red boxes. For instance, the y-axis slice on the bottom (1957-1958) compares from 0 forward up to +43 years (1999-2000), while the slice on the top (1999-2000) compares backward from 0 up to -43 years (1957-1958).}
    \label{fig:asynchronous_summed_kld}
\end{figure}

In addition to the synchronous comparison, we calculate total divergence for each time slice against all other time slices, both past and future. This asynchronous comparison helps identify time slices with the least and most divergence from each other. As illustrated in Figures \ref{fig:asynchronous_summed_kld}, comparing Silk and Treder again reveals notable differences. 

From the 1970s onward, Treder’s overall KLD values decrease toward earlier years, suggesting his terminology aligns more closely with past mainstream terminology. For instance, in 1999-2000, Treder’s terminology is most similar to that of the mainstream in 1957-1958. Conversely, Silk’s terminology, with few exceptions, aligns with terms used by the mainstream in the future. This may suggest that Silk or the topics he writes about are influential on the future direction of GRG research. While we cannot deduce direct influence from this comparison alone, citation numbers (see \ref{subsec:General statistics for both authors}) support this interpretation. On the other hand, Treder's results align with the above mentioned findings: his research seems to focus on foundational questions in physics, often already raised in earlier years of GRG research. The references he cites and the high KLD values for names he uses, which are from the late 19th and early 20th century like Einstein, Mach, Schrödinger or others, further indicate this focus. 

\newpage
\subsection{Method 2 - Changing Densities}\label{subsec:Method 2 - Changing Densities}

We will now turn to our second method. First, we transform all publications into embeddings vectors and estimate density as outlined in \ref{subsubsec:Method 2 - Density approach}. We then plot EDE over time for each publication of the respective individual.   

\subsubsection{Synchronous Comparison}\label{subsubsec:Synchronous Comparison}

\begin{figure}[ht!]
    \centering
    \begin{subfigure}[b]{\textwidth}
        \includegraphics[width=\linewidth]{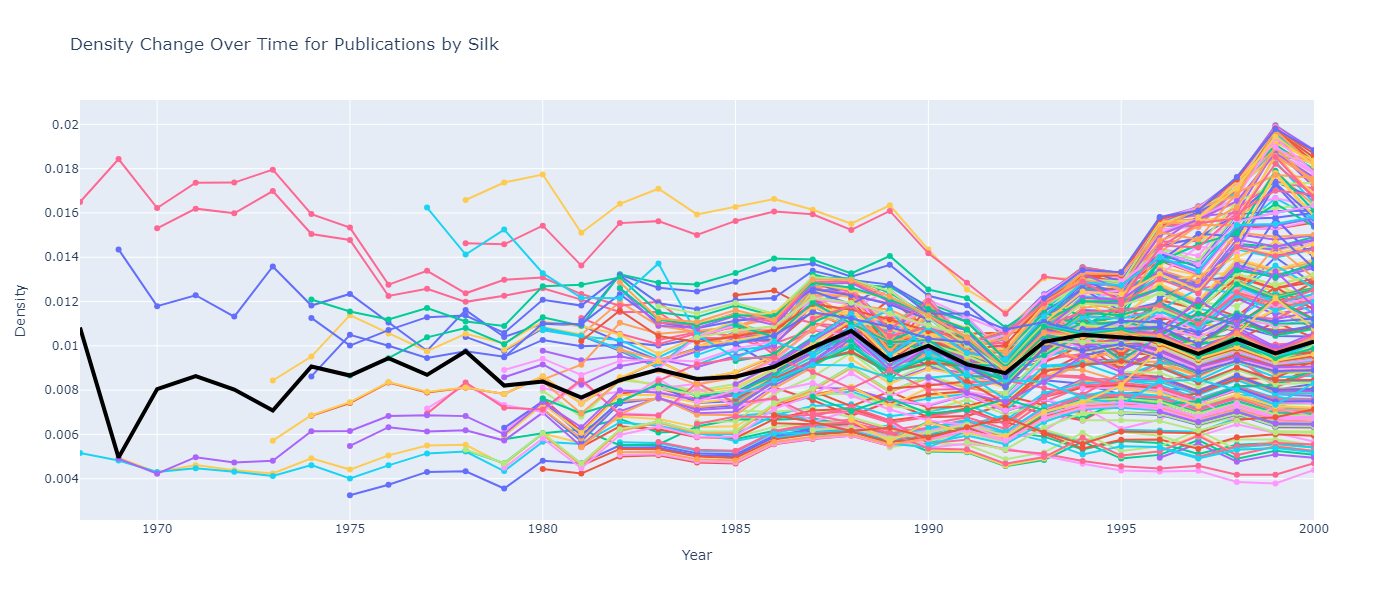}
    \end{subfigure}
    \begin{subfigure}[b]{\textwidth}
        \includegraphics[width=\linewidth]{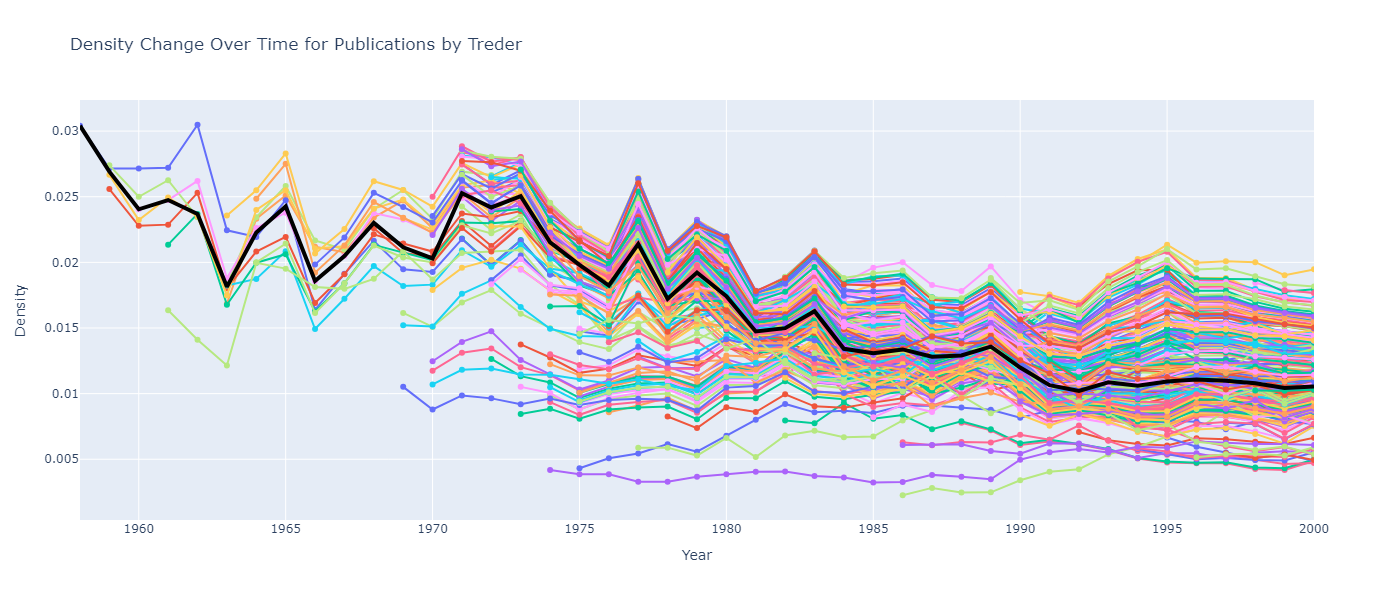}
    \end{subfigure}
    \caption{Embeddings Density Estimation (EDE) over time for publications by Silk (top) and Treder (bottom), showing shifts in density around their publications across time slices. The thick black line represents the median value of all publications, indicating the general trend.}
    \label{fig:density_change_over_time}
\end{figure}

In Figure \ref{fig:density_change_over_time}, EDE is plotted over time for all publications by Silk and Treder, overlaying each other. Both trajectories reveal a pattern consistent with our analysis in \ref{subsec:Method 1 - Changing Frequencies}. 

During the 1960s to mid-1970s, we again see a fluctuating pattern for both, indicative of the formation period. But starting with the mid-1970s, where Silk’s KLD values stabilise and remain consistently low, the median EDE also stabilises and slightly increases. This suggests that an increasing number of publications in the whole field appear in the vicinity of Silk’s work, meaning that the proportion of newly published papers similar to his research is growing. A comparable but inverse trend is observed for Treder. From the mid- to late-1970s, as his KLD values start to rise, his mean EDE starts to decline, indicating that the proportion of new publications that are clustering around the topics he is writing about is declining.

\section{Conclusion}\label{sec:Conclusion}  
We presented an approach to trace individual knowledge trajectories within the formation of a scientific field by combining information-theoretic measures with density estimation of sentence embeddings.

Our analysis from 1957 to 2000 on Joseph Silk and Hans-Jürgen Treder shows that Treder increasingly uses terminology associated with the \enquote{past} of the mainstream. This shift correlates with a decline in EDE around his publications, indicating relative decreasing research activity on his topics correlating with a lower citation rate. In the late 1950s to early 1970s, Treder’s focus on fundamental principles positioned him within the mainstream of General Relativity and Gravitation (GRG) research. Treder’s trajectory diverges, as GRG research takes a turn toward astrophysics: 

\blockquote[{\cite[p. 541-542]{blum_gravitational_2018}}]{On the theoretical side, the 1960s brought a radical refocusing of the research agendas of many general relativity experts. Following the discovery of quasars in 1963 and other important astronomical discoveries, the interest of the experts switched towards the newly established fields of relativistic astrophysics and observational cosmology, moving away from the 1950s research agendas still related to classical unified field theories or quantum theories of gravity.}

In contrast, Joseph Silk, one of the most cited authors in the corpus, increasingly uses \enquote{future-oriented} terminology, suggesting he shapes GRG terminology. Relative research activity around Silk’s topics increases over time, reflecting his role in the above described \textit{astrophysical turn}. 

Our analysis therefore seems to support the assumption given in \ref{sec:Analysis}, that KLD and EDE provide complementary insights. KLD measures divergence from mainstream language, while EDE indicates how closely new publications cluster around an author’s work. Together, they enable quantitative analysis of individual versus system knowledge evolution. Lower KLD suggests an author's terminology aligns with mainstream language, correlating with higher EDE, meaning other researchers frequently engage with similar topics. Whether this pattern holds for all cases remains to be tested, but here our aim was to combine a quantitative view of global structural changes with local case studies rather than to establish a general rule. 

Future steps include integrating citation analysis to explore if individuals whose terminology aligns with the mainstream — and who experience rising density — also show increase in citation rates. This would evaluate our findings from the seen citation trends in \ref{subsec:General statistics for both authors}. Full-text data, beyond titles and abstracts, would also strengthen our results, as would a larger and less biased sample. As mentioned earlier, collection biases within the dataset, drawn from relatively small samples, remain mostly unquantified, which is a critical consideration for structural analysis. Additionally, the sparsity and length of texts is another limitation. In this dataset, we only had access to titles and abstracts, and sometimes only titles. Therefore, a comparison of individuals with relatively few publications would not be particularly meaningful (in contrast to the two cases we selected with many publications). With full-text data, the approach would yield more robust results. The inclusion of classical citation networks could allow to track the development of the reception of individuals within the global setting. This would potentially answer the question, if there is a correlation between all three measures: high EDE, mainstream language (low KLD) and high citation count or vice versa. 

Finally, we here used the presented approach for the tracking of individual knowledge trajectories of physicists within the field of GRG. But the approach is transferable to other fields of science and can also be applied in settings outside the highly formalised scenarios of scientific publications, such as knowledge dissemination in migration processes or the transition of knowledge from science into policy.


\subsection*{Acknowledgements}

The authors acknowledge funding from the Federal Ministry of Education and Research (ModelSEN, Grant No. 01UG2131) at the Max Planck Institute for the History of Science. We thank the members of the ModelSEN team for their valuable feedback, with special thanks to Aleksandra Kaye and Bernardo S. Buarque.


\begingroup
\small 
\bibliography{references}
\endgroup
\end{document}